\documentclass[lettersize,journal]{IEEEtran}
\usepackage{amsmath,amsfonts}
\usepackage{algorithmic}
\usepackage{algorithm}
\usepackage{array}
\usepackage[caption=false,font=normalsize,labelfont=sf,textfont=sf]{subfig}
\usepackage{textcomp}
\usepackage{stfloats}
\usepackage{url}
\usepackage{verbatim}
\usepackage{graphicx}
\usepackage{cite}
\usepackage{balance}

\usepackage{algorithm}
\usepackage{algorithmic}

\usepackage[utf8]{inputenc} 
\usepackage[T1]{fontenc}    
\usepackage[colorlinks=true, linkcolor=red,
citecolor=blue,urlcolor=blue]{hyperref}       
\usepackage{url}            
\usepackage{booktabs}       
\usepackage{amsfonts}       
\usepackage{nicefrac}       
\usepackage{microtype}      
\usepackage{bbding}
\usepackage{graphicx}
\usepackage{color}
\usepackage{amsmath,amssymb,amsfonts}
\usepackage{graphicx}
\usepackage{textcomp}
\usepackage{bm}
\usepackage{multirow}
\usepackage{booktabs}
\usepackage{bbding}
\usepackage{mfirstuc}
\usepackage[misc]{ifsym} 
\usepackage{graphics}
\usepackage{makecell}
\usepackage{amsthm}
\usepackage{color}

\newcommand{\tabincell}[2]{\begin{tabular}{@{}#1@{}}#2\end{tabular}} 

\newcommand{\gbf}[1]{\bf{ (#1)}}
\newcommand{\rbf}[1]{\bf{ (#1)}}

\newtheorem{theorem}{Theorem}

\usepackage{xspace}
\newcommand{\vct}[1]{\boldsymbol{#1}} 
\newcommand{\mat}[1]{\boldsymbol{#1}} 
\newcommand{\ie}{\emph{i.e.}\xspace} 
 
\newcommand{\etc}{\emph{etc}\xspace} 
\newcommand{\etal}{\emph{et al.}\xspace}

\newcommand{\KL}[2]{\mathrm{KL}[#1\|#2]}

\newcommand{\methodname}{CORE\xspace}
\newtheorem{proposition}[theorem]{Proposition}
\newtheorem{definition}[theorem]{Definition}

\newlength\savewidth\newcommand\shline{\noalign{\global\savewidth\arrayrulewidth
  \global\arrayrulewidth 1.5pt}\hline\noalign{\global\arrayrulewidth\savewidth}}

\begin{document}

\title{\methodname: Learning Consistent Ordinal REpresentations for Image Ordinal Estimation}

\author{Yiming Lei,~\IEEEmembership{Member,~IEEE}, Zilong Li, Yangyang Li, Junping Zhang,~\IEEEmembership{Senior Member,~IEEE},\\ and Hongming Shan,~\IEEEmembership{Senior Member,~IEEE}
\thanks{Y. Lei, Z. Li and J. Zhang are with the Shanghai Key Laboratory of Intelligent Information Processing, the School of Computer Science, Fudan University, Shanghai 200433, China (e-mail: ymlei@fudan.edu.cn; zilongli21@m.fudan.edu.cn; jpzhang@fudan.edu.cn).}
\thanks{Y. Li is with the Academy of Mathematics and Systems Science, Chinese Academy of Sciences, Beijing 100190, China (e-mail: yyli@amss.ac.cn).}
\thanks{H. Shan is with the Institute of Science and Technology for Brain-Inspired Intelligence and MOE Frontiers Center for Brain Science, Fudan University, Shanghai 200433, and also with the Shanghai Center for Brain Science and Brain-Inspired Technology, Shanghai 200031, China (e-mail: hmshan@fudan.edu.cn).}

}

\markboth{}%
{Shell \MakeLowercase{\textit{et al.}}: A Sample Article Using IEEEtran.cls for IEEE Journals}


\maketitle


\begin{abstract}
The goal of image ordinal estimation is to estimate the ordinal label of a given image with a  convolutional neural network. 
Existing methods are mainly based on ordinal regression and particularly focus on modeling the ordinal mapping from the feature representation of the input to the ordinal label space. 
However, the manifold of the resultant feature representations does not maintain the intrinsic ordinal relations of interest, which hinders the effectiveness of the image ordinal estimation.
Therefore, this paper proposes learning intrinsic Consistent Ordinal REpresentations (\methodname) from ordinal relations residing in ground-truth labels while encouraging the feature representations to embody the ordinal low-dimensional manifold. 
First, we develop an ordinal totally ordered set (\emph{toset}) distribution (OTD), which can (i) model the label embeddings to inherit ordinal information and measure distances between ordered labels of samples in a neighborhood, and (ii) model the feature embeddings to infer numerical magnitude with unknown ordinal information among the features of different samples. 
Second, through OTD, we convert the feature representations and labels into the same embedding space for better alignment, and then compute the Kullback‒Leibler (KL)-divergence between the ordinal labels and feature representations to endow the latent space with consistent ordinal relations. 
Third, due to the covariate shift from the feature representations to ordinal labels, we optimize the KL-divergence through ordinal prototype-constrained convex programming with dual decomposition; our theoretical analysis shows that we can obtain the optimal solutions via gradient back-propagation. 
The results show that the proposed \methodname is easily implemented and plug-and-play compatible with existing methods without needing to modify network architectures.
Extensive experimental results on four different scenarios---facial age estimation, medical disease progression prediction, historical image dating prediction, and image aesthetics assessment---demonstrate that the proposed \methodname can accurately construct an ordinal latent space and significantly enhance existing deep ordinal regression methods to achieve new state-of-the-art results.
\end{abstract}



\section{Introduction}

\IEEEPARstart{T}{he} goal of ordinal estimation is to estimate the ordinal label of a given image with a convolutional neural networks (CNN), which is often formulated as ordinal regression (OR),  a classic problem in machine learning tailored for making predictions for images with ordered labels. 
Typical applications include facial age estimation from \emph{young} and \emph{middle age} to \emph{old}, movie ratings from \emph{one star} to \emph{five stars}, and disease progression ratings from \emph{healthy} and \emph{mild} to \emph{severe}. 
Due to its use of ordered labels, ordinal regression is an intermediate problem between classification and regression~\cite{CORFs,LDLF}.

\begin{figure}[t]
\centerline{\includegraphics[width=1.0\linewidth]{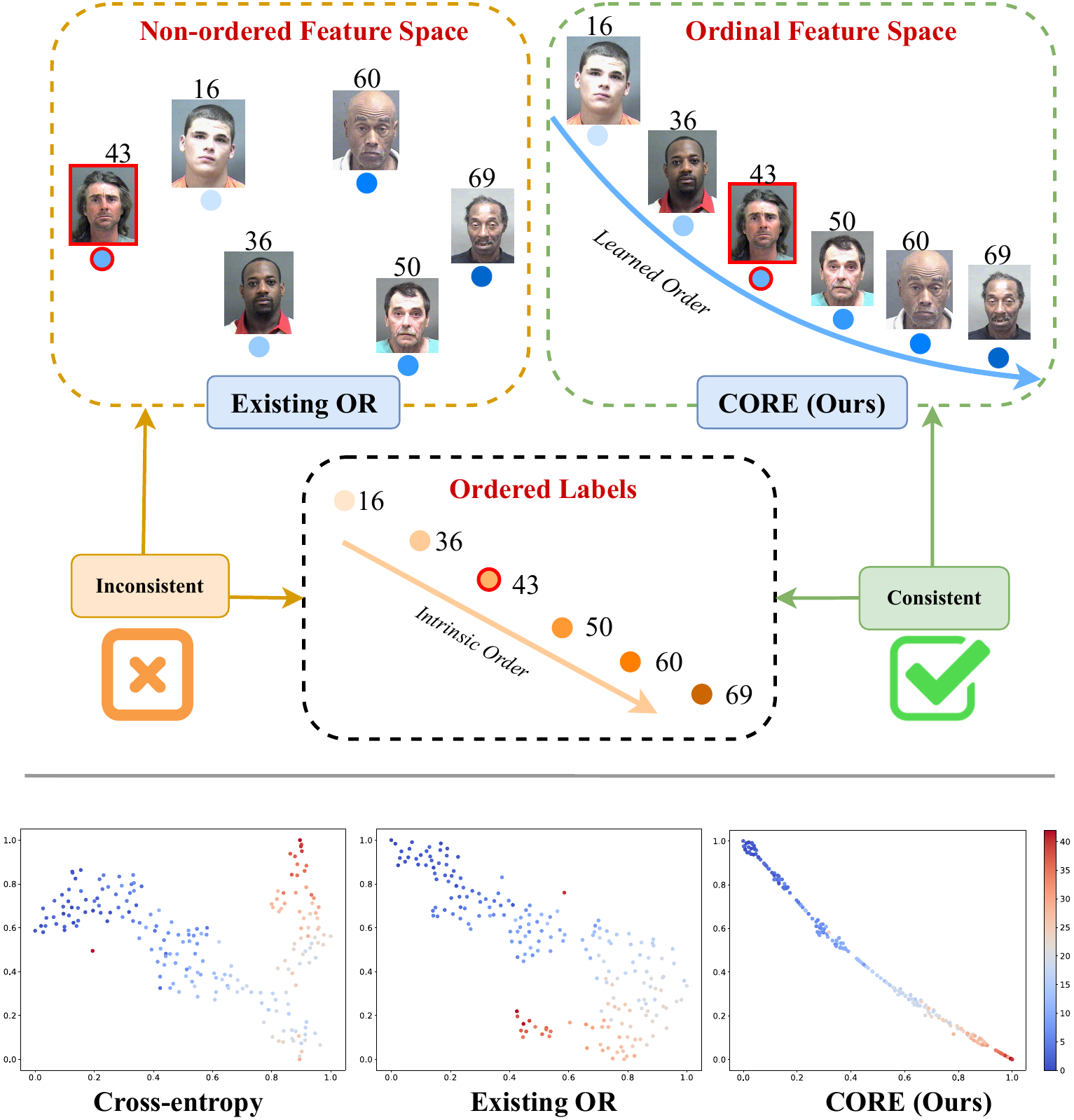}}
\caption{Comparison between existing methods and our \methodname. (\textbf{Top}) Existing ordinal regression methods yield feature representations with no ordered low-dimensional manifold. In contrast, the goal of \methodname is to align the feature representations explicitly with the ordered labels via KL-divergence, then the feature space exhibits a nearly one-dimensional ordinal manifold. (\textbf{Bottom}) The t-SNE visualizations on the MORPH dataset using different OR methods.
}
\label{fig:motivation}
\end{figure}

In the past few years, developments in deep neural networks have led to similar advancements in the development of ordinal regression methods~\cite{DRFs,CORFs,Mean-variance,POE,SORD}, which have subsequently come to be  referred to as deep ordinal regression. 

These methods mainly focus on modeling an ordinal mapping from the feature representation space of the input to the ordinal label space. 
For example, a popular way to model ordinal information is to formulate ordinal regression as a multibinary classification problem since the ordinal labels are monotonically distributed~\cite{DRFs,CORFs,niu2016ordinal,fu2018deep,LDLF}. These methods tend to identify whether the prediction of an output unit is greater than a rank, and the final prediction is the sum of the predictions of all binary classifiers. 
Another kind of method aims to construct distributions that can preserve ordinal information for the logits of convolutional neural networks or ordinal labels. Poisson, binomial, and cumulative probabilities of multibinary classifiers are imposed on output logits and work well in some domains such as medical image classification~\cite{Unimodal,NSB}. Soft ORDinal vectors (SORD) and Beta distribution are also applied to convert the label space~\cite{SORD,Beta}.
Recently, the frameworks based on a combination of CNNs and random forest have been proposed for learning the ensemble of feature partitions at splitting nodes and ordinal distributions at leaf nodes~\cite{LDLF,DRFs,CORFs,lei2022meta}; these frameworks can be trained using end-to-end back-propagation. 
However, these methods do not consider refining the distribution of feature representations in the latent space, which should, as expected, be equipped with an ordinal manifold consistent with that of ordinal labels. As shown at the bottom of Fig.~\ref{fig:motivation}, the features learned by cross-entropy (CE) are distributed in a less ordered manner. Although the existing OR methods can rectify the feature distribution to exhibit ordinal relations across different classes, they still cannot approximate the intrinsic order of the ground-truth labels, \ie, their nearly one-dimensional order, as shown at the top of Fig.~\ref{fig:motivation}. 

In this work, we address ordinal regression from a new perspective that is \emph{orthogonal} to the current research direction and compatible with existing deep ordinal regression methods. We propose learning consistent ordinal representations in embedding space, termed~\methodname, which enables the trained models to preserve the low-dimensional ordinal structures of feature representations, as illustrated at the bottom of Fig.~\ref{fig:motivation}. Unlike the current deep ordinal regression, which focuses on learning a mapping from the input to the feature and then to the ordinal label space, \methodname focuses on maintaining ordinal consistency between feature representations and ordinal labels. We are motivated by the fact that although there is ordinal-unrelated information in parts of input images such as the background, the geometrical structure of the features in the latent space should be dominated by ordinal information provided by ordered labels.
To this end, we propose aligning the feature representations with ordinal labels in the same embedding space.
Here, we consider the relations between one sample and other samples in a mini-batch, and then we regard their corresponding ordinal labels as a totally ordered set (\emph{toset})~\cite{bourbaki2004theory}, which maintains ordinal relations among all the elements. Accordingly, we propose the ordinal \textit{toset} distribution (OTD) to model a set of ordinal labels as a probability distribution, \ie, label embedding, conserving ordinal information. Similarly, the feature representations can also be modeled as a probability distribution with the same dimension as the label embedding, \ie, feature embedding. Consequently, our goal is to encourage the feature embeddings to learn ordinal information preserved by label embeddings by minimizing the  Kullback-Leibler (KL) divergence between them.

Furthermore, instead of directly minimizing the KL-divergence, we introduce an ordinal prototype constraint that enforces intraclass compactness and interclass continuity within a mini-batch. Hence, we propose to integrate this constrained convex optimization program into end-to-end back-propagation~\cite{amos2017optnet,agrawal2019differentiable,berthet2020learning,ron2022dual}. Considering that the label embeddings and feature embeddings are derived from different domains suffering from internal covariate shift, we infer the convex dual-decomposition of the objective KL-divergence with two assumed distributions to alleviate the influence of the covariate shift. 
Extensive experimental results on four different applications demonstrate the effectiveness of the proposed \methodname when coupled with existing deep ordinal regression methods.

The contributions of this work are summarized as follows.
\begin{itemize}
    \item We propose learning consistent ordinal representations (\methodname) for image ordinal estimation, regularizing the feature representations with ordinal constraints. To the best of our knowledge, this is the first attempt to learn consistent ordinal representation, which is orthogonal to the current research direction.
    \item We develop an ordinal \emph{toset}  distribution (OTD), which can project the features and labels into the same embedding space while maintaining the ordinal relations across ranks. 
    \item We introduce an ordinal prototype-constrained convex
programming with dual decomposition to optimize the KL-divergence to avoid the covariate shift from the feature representations to ordinal labels; our theoretical analysis shows that we can obtain the optimal solutions via gradient back-propagation. 
    \item \methodname is plug-and-play compatible with existing deep ordinal regression methods and requires no modifications to their architectures. Extensive experiments on four different applications demonstrate the superiority of the proposed \methodname in terms of quantitative and qualitative results. Visualization of the  feature representations further confirms the dominance of ordinal information in the latent space.
\end{itemize}

The remainder of this paper is organized as follows. Section~\ref{sec:related} briefly reviews existing works on ordinal regression for image ordinal estimation and convex optimization with deep neural networks. In Section~\ref{sec:method}, we present the proposed \methodname and its dual optimization. Section~\ref{sec:experiment} provides extensive experimental results on four applications and demonstrates the state-of-the-art performance of the proposed method. Finally, we summarize the paper in Section~\ref{sec:conclusion}.


\section{Related Work}
\label{sec:related}
In this section, we briefly review the existing works on topics including ordinal regression with traditional solutions and deep learning-based methods as well as convex optimization integrated into deep neural networks.

\subsection{Ordinal Regression for Image Ordinal Estimation} Image ordinal estimation is often solved by 
ordinal regression methods, which can be grouped into three categories: na\"{i}ve methods, ordinal binary decomposition, and threshold methods~\cite{survey2016}. Na\"{i}ve methods include standard regression techniques such as support vector regression~\cite{svr} and nominal classification~\cite{multisvm}. Ordinal binary decomposition has been proposed for independently solving multiple binary classification problems~\cite{frank2001simple,cardoso2007learning,waegeman2009ensemble,tip2015} since the ordinal information provides the possibility of comparing the different ranks. Threshold-based methods are used to estimate the intervals of different labels under the assumption that there exists a continuous latent variable that underlies the intrinsic ordinal information~\cite{verwaeren2012learning,tip2018}. We note that the proposed \methodname learns the intrinsic ordinal manifold in the latent space, which is also consistent with this assumption but places greater emphasis on the consistency between the feature representations and ordinal labels.

Based on deep learning techniques, OR achieves further performance improvements, and its corresponding methods are also related to those mentioned above. The main deep OR methods recently proposed in the literature can be grouped into three categories. 
\textit{(i)} Multi-binary classification methods tend to identify whether the prediction of an output unit is greater than a rank, and the final prediction is the sum of the predictions of all binary classifiers~\cite{niu2016ordinal,fu2018deep}. 
\textit{(ii)} Architectures combining CNN and random forest can be used to learn an ensemble of feature partitions at split nodes and ordinal distributions at leaf nodes. Deep regression forests (DRFs)~\cite{DRFs} validate the ability to simultaneously back-propagate the gradients of the integration of CNNs with random forests. This method inspired the creation of label distribution learning forests~\cite{LDLF} and convolutional ordinal regression forests (CORFs)~\cite{CORFs} to further leverage this flexibility to improve the ordinal regression for facial age estimation. In addition, the meta ordinal regression forest (MORF) solves the two problems of CORFs that hinder performance improvement, \ie, the fixed structure of the forest and the large variances among predictions of independent trees, and outperforms CORFs in medical image progression estimation~\cite{lei2022meta}.
\textit{(iii)} Distribution-based models aim to construct distributions that preserve ordinal information for CNN output logits or ordinal labels. Poisson, binomial, and cumulative probabilities of multi-binary classifiers are imposed on output logits, and the models worked well in medical image classification. Soft ORDinal vectors (SORD) and Beta distributions are applied to impose unimodal distribution on label space, but they maintain the same converted label for samples of the same rank while not taking neighbor relations into account~\cite{SORD,Beta}.

However, these methods lack the consideration of constraining feature representations that contain too much information unrelated to the ordinal information. The proposed \methodname corrects the ordinal information in the latent space, is \emph{orthogonal} to existing works and is compatible with existing deep ordinal regression methods.

\subsection{Convex Optimization with Deep Neural Networks}
Convex optimization has been introduced into deep learning architectures via carefully designed, translated formulation and differentiable mapping layers. 
Montavon~\etal introduced OptNet, the first successful method that computes the backward gradients of quadratic programs in an end-to-end manner~\cite{montavon2017explaining}.  Agrawal~\etal proposed disciplined parametrized programming for convex optimization while differentiating through back-propagation~\cite{agrawal2019differentiable}. Moreover, some non-differentiable operations integrated with sampling and sequential computations have also been explored~\cite{domke2010implicit,paulus2020gradient,berthet2020learning}; however, they also cause heavy computational overhead. Dual decomposition is a classical technique for converting primal optimization problems in machine learning~\cite{rush2012tutorial,sra2012optimization}, providing additional power to deep neural networks with successful applicability. A learnable consistent attention layer was proposed in~\cite{ron2022dual}, which back-propagates the gradients of the decomposed distribution alignment problem via strong duality. 
Purica~\etal proposed a convex optimization framework using a dual splitting-based variational formulation to enhance video quality and resolution~\cite{purica2018convex}. Takeyama and Ono adopted a hybrid spatial-spectral total variation constraint for robust hyperspectral image fusion, and the convex problem was solved by a primal-dual splitting algorithm~\cite{takeyama2022robust}. Different from the above methods, our \methodname aim to bridge embedding vectors of different modalities in latent space, \ie, images and ordered labels. To achieve this goal, we solve an ordinal prototype-constrained convex problem using dual decomposition, and the theoretical results guarantee the optimal solutions.


\section{Methodology}
\label{sec:method}

\begin{figure*}[ht]
\centerline{\includegraphics[width=1.0\linewidth]{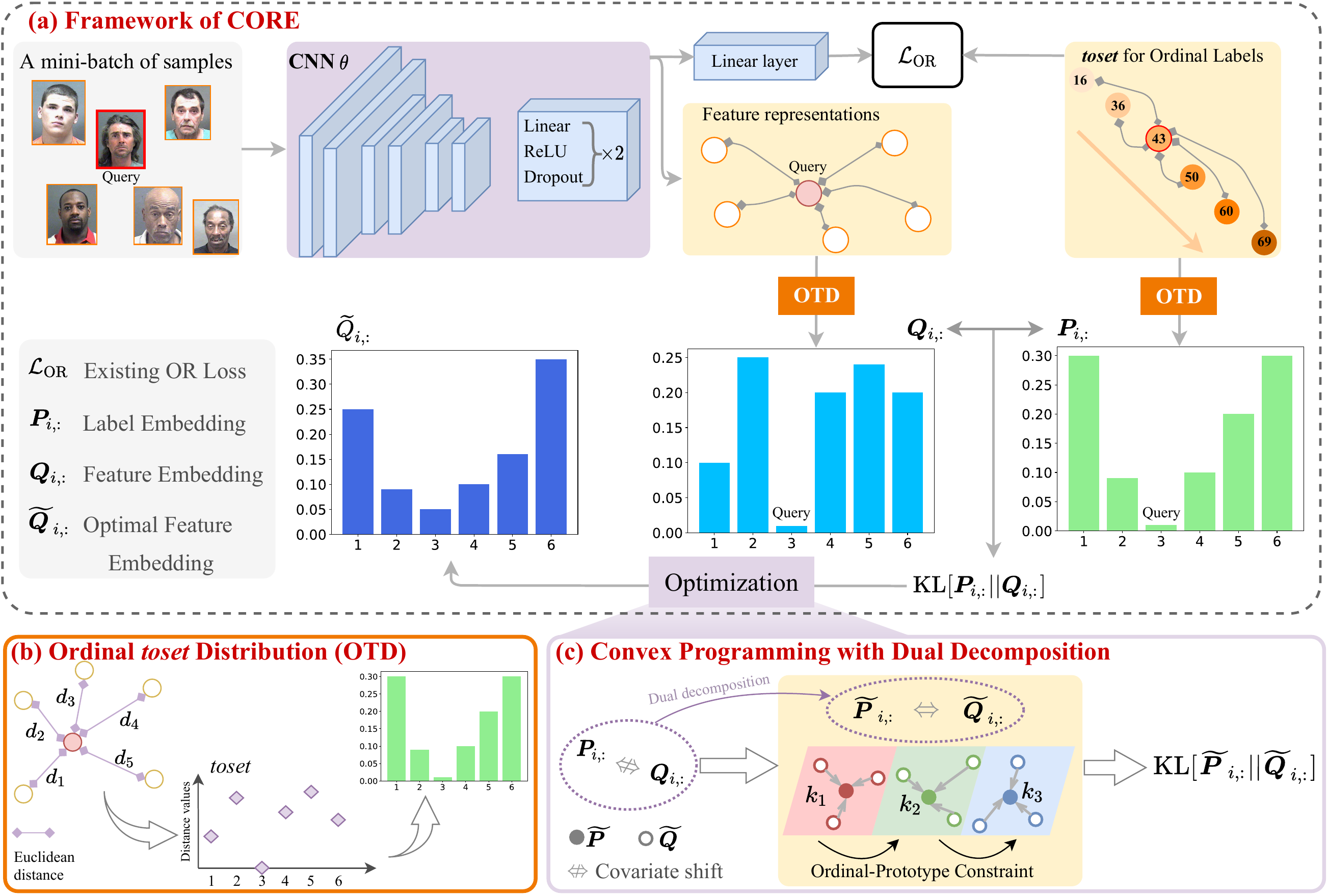}}
\caption{Main framework of the proposed \methodname. (a) A mini-batch of samples are mapped to latent space through a CNN ($\mat{\theta}$). Then, these feature representations and corresponding ordinal labels are converted to probabilities via (b) OTD, \ie, $\mat{Q}_{i, :}$ and $\mat{P}_{i, :}$, respectively. Hence, (c) the consistent ordinal representation learning is conducted between $\mat{P}$ and $\mat{Q}$ via dual decomposition coupled with ordinal-prototype constraint.}
\label{fig:framework}
\end{figure*}

Fig.~\ref{fig:framework} presents the main framework of the proposed \methodname. In the following, we first introduce the formulation for the problem of image ordinal estimation. Then we present the proposed ordinal OTD as the embedding of the ordered labels that inherits ordinal information. Next, based on the OTD, we further describe how to encourage consistent ordinal representation learning between feature representations and ordered labels in the embedding space. Finally, we derive a dual decomposition for the ordinal prototype-constrained optimization and provide the corresponding theoretical analysis of the optimal solutions.

\subsection{Image Ordinal Estimation}
Let $\mathcal{X}$ be the image space and $\mathcal{Y}=\{r_{1}, \ldots, r_{C}\}$ the label space, where $r_{1} \preceq r_{2} \preceq \ldots \preceq r_{C}$, and $\preceq$ denotes the order among the ordinal ranks. The image ordinal estimation learns the mapping: $h: \mathcal{X} \to \mathcal{Y}$. In the context of deep learning, when given a set of $N$ observations $\{(\mat{x}_{i}, {y}_{i})\}_{i=1}^{N}$ drawn \emph{i.i.d.} from $\mathcal{X} \times \mathcal{Y}$, where $y_{i} \in \mathcal{Y}$, a simple neural network can be trained end-to-end, resulting in an ordinal estimation result. Specifically, this process can usually be regarded as the combination of feature representation learning, $\vct{z}=f_{\vct{\theta}}(\vct{x})$, and ordinal label mapping, ${y}=g_{\vct{\phi}}(\vct{z})$; here, $f$ and $g$ are the backbone network, parameterized with $\mat{\theta}$, and the linear layer, parameterized with $\mat{\phi}$, respectively.

The existing deep image ordinal estimation is a discriminative model that learns a conditional probability distribution $p(y|\mat{x}) = g_{\mat{\phi}}(f_{\mat{\theta}}(\mat{x}))$. It also forms a Markov chain
\begin{align}
\mat{x} \to \mat{z} = f_{\mat{\theta}}(\mat{x})\to y = g_{\mat{\phi}}(\mat{z}),
\label{eq:markov}
\end{align}
where $\mat{z}$ is the feature representation of the input $\mat{x}$. Intuitively, $\mat{z}$ contains all the discriminative information related to different ordered labels. From the perspective of manifold learning, $y$ can be regarded as the low-dimensional representation of input $\mat{x}$ that is embedded in a high-dimensional space~\cite{tenenbaum2000global}. That is, $\mat{z}=f_{\mat{\theta}}(\mat{x})$ implies the ordinal information corresponding to that in $y$.

Unfortunately, if the ordered labels only serve as supervisions such as those in conventional classification and regression strategies, the learning pipeline in Eq.~\eqref{eq:markov} has difficulty enabling representation $\mat{z}$ to inherit ordinal relations as follows~\cite{kim2020ordinal}:
\begin{align}
||\mat{z}_{1} - \mat{z}_{3}|| \geq \max\Big\{||\mat{z}_{1} - \mat{z}_{2}||, ||\mat{z}_{2} - \mat{z}_{3}||\Big\},
\label{eq:or_constraint}
\end{align}
where $\mat{z}_{1}, \mat{z}_{2}, \mat{z}_{3}$ are feature vectors corresponding to labels $y_{1} \preceq y_{2} \preceq y_{3}$. In this work, we propose an OTD to model the representation $\mat{z}$ and intrinsic ordinal information residing among labels in $\mathcal{Y}$. Hence, we can bridge them in an embedding space and preserve the ordinal relations for feature representation as described in Eq.~\eqref{eq:or_constraint}.

\subsection{Ordinal \textit{toset} Distributions}
In this paper, our aim is to encourage the feature representation $\vct{z}$ to inherit ordinal relations in $\mathcal{Y}$. Intuitively, we can only approximate the one-dimensional ordinal information present among the ordered labels. However, for each of the $N$ observations $\{\mat{x}_{i}, y_{i}\}$, label $y_{i}$ is independent of the others. Therefore, we propose a label embedding that implies ordinal relations between $y_{i}$ and $y_{j}, j\neq i$. 
Inspired by~\cite{kim2020ordinal,liu2022ordinal}, we intend to impose the ordinal constraint in Eq.~\eqref{eq:or_constraint} on the label embedding so that all the elements in this embedding vector should satisfy the properties of a \emph{totally ordered set}.

\begin{definition}[\cite{bourbaki2004theory}] The totally ordered set (\emph{toset}) is a set of $N$ elements, $\mathcal{O} = \{o_{i}\}_{i=1}^{N}$, equipped with a total order relation ``$\preceq$'', satisfying:
\begin{itemize}
\item Reflexivity: $o_{i} \preceq o_{i}, \forall o_{i} \in \mathcal{O}$.
\item Antisymmetry: if $o_{i} \preceq o_{j}$ and $o_{j} \preceq o_{i}$, then $o_{i} = o_{j}$.
\item Transitivity: if $o_{i} \preceq o_{j}$ and  $o_{j} \preceq o_{l}$, then $o_{i} \preceq o_{l}$.
\item Comparability: $\forall o_{i}, o_{j} \in \mathcal{O}$, either $o_{i} \preceq o_{j}$ or $o_{j} \preceq o_{i}$.
\end{itemize}
\label{defnition:toset}
\end{definition}

First, we can compute the inter-class distances between $y_{i}$ and other naturally ordered classes~\cite{SORD}. Note that this kind of distance is meaningless in traditional classification problems such as identifying incomparable pairs  ``\textit{cat}'' and ``\textit{dog}'', which contradicts the property of \emph{Comparability} in Definition~\ref{defnition:toset}.

Second, we let $d(\cdot, \cdot)$ be the Euclidean distance. Then, the distance between the labels of two samples is calculated as $d(y_{i}, y_{j})$. For each observation $\{(\mat{x}_{i}, {y}_{i})\}$, we construct a set $\mathcal{P}_{i}$ for label $y_{i}$ that describes the distances between $y_{i}$ and $y_{j}$: $\mathcal{P}_{i} = \{ d(y_{i}, y_{j}) \}_{j=1}^{N}$, where $i = 1, 2, \ldots, N, j \neq i$, and $d(y_{i}, y_{i}) = 0$. Based on Definition~\ref{defnition:toset}, we make the following proposition to show that each $\mathcal{P}_{i}$ is a \textit{toset} that inherits the ordinal constraint in Eq.~\eqref{eq:or_constraint}.
\begin{proposition}
The label set $\mathcal{Y}$ in ordinal regression is a \emph{toset} that satisfies ordinal constraint in Eq.~\eqref{eq:or_constraint}. Furthermore, we calculate the Euclidean distances between any pairs $(y_{i}, y_{j})$ as $d(y_{i}, y_{j})$, then we can construct an ordinal distance set for the $i$-th sample as $\mathcal{P}_{i} = \{d(y_{i}, y_{j})\}_{j=1}^{N}, i = 1, \dots, N$, where $d(y_{i}, y_{i}) = 0$, then, $\forall y_{i}$, $\mathcal{P}_{i}$ is also a \emph{toset} and satisfies the ordinal constraint.
\label{proposition:toset}
\end{proposition}

Different from the recursively conditional Gaussian set in~\cite{liu2022ordinal} which reorders the latent units, each element in $\mathcal{P}_{i}$ explicitly represents the distance between two classes. 
Based on Proposition~\ref{proposition:toset}, we define a matrix $\mat{P} \in \mathbb{R}^{N \times N}$, in which each row vector $\mat{P}_{i, :} \in \mathbb{R}^{1 \times N}$ denotes the embedding (OTD) for label $y_{i}$ and is composed of the following pairwise distances:
\begin{definition} Ordinal toset distribution (OTD) is a normalized vector that each element $\mat{P}_{i, j}$ denotes the similarity between label $y_{i}$ and $y_{j}$, where $d(y_{i}, y_{i}) = 0$:
\begin{align}
\mat{P}_{i,j} = \frac{\exp\left(- d(y_{i}, y_{j}) \right)}{\sum\limits_{\substack{n=1\\n\neq i}}^N \exp\big(- d(y_{i}, y_{n}) \big)},\quad  j \in \{1, \ldots, N\}.
\label{eq:p_Y}
\end{align}
\label{def:otd}
\end{definition}

Therefore, all the corresponding labels $y_{i}$ can be converted to $\mat{P}_{i,:} \in \mathbb{R}^{1 \times N}$, and each dimension in $\mat{P}_{i, :}$ represents the probability that label $y_{i}$ picks $y_{j}$ ($j \neq i$) as its neighbor~\cite{tsne}. This probability is similar to the soft labels in SORD~\cite{SORD}; nevertheless, we clarify the differences as follows: SORD assigns all samples of the same class with identical Gaussian-like distributions that are generated by the fixed label set, while the proposed OTD focuses on a number of samples, \ie, the OTD of a given sample varies according to its neighbors. Under the training procedure for randomly sampled mini-batches of the entire set of samples, the OTD can encourage target samples to learn diverse relations for different training iterations, which will be discussed in the next section. However, SORD does not enjoy this advantage.

\subsection{Consistent Ordinal REpresentation (\methodname) Learning}
Existing deep ordinal regression methods do not  consider \emph{explicitly} imposing an ordinal constraint on latent feature representations. In this work, we encourage $\vct{z} = f_{\vct{\theta}}(\vct{x})$ to have an ordinal constraint in Eq.~\eqref{eq:or_constraint}. For the supervised learning framework, we have no prior ordinal information on feature representations $\vct{z}_{i}$. For a set of samples with ordinal labels, we can only measure the Euclidean distances between the  feature representations of different samples in latent space, \ie, $d(\vct{z}_{i}, \vct{z}_{j})$. Similar to \emph{toset} $\mathcal{P}_{i}$, we also construct a corresponding set for feature representation $\vct{z}_{i}$: $\mathcal{Q}_{i} = \{d(\vct{z}_{i}, \vct{z}_{j})\}_{j=1}^N, i=1, \dots, N$, $d(\vct{z}_{i}, \vct{z}_{i}) = 0$. Obviously, $\mathcal{Q}_{i}$ only reflects the numerical magnitude of the distances between $\vct{z}_{i}$ and $\vct{z}_{j}$. Therefore, according to Proposition~\ref{proposition:toset}, we cannot guarantee that $\mathcal{Q}_{i}$ is a \textit{toset} nor that it possesses the property of ordinal constraint in Eq.~\eqref{eq:or_constraint}.

Therefore, we expect the feature representations to learn ordinal relations in $\mathcal{P}_{i}$. Specifically, we define a matrix $\mat{Q} \in \mathbb{R}^{N \times N}$ in which each row vector $\mat{Q}_{i, :}$ is a distribution, and each element represents the pair-wise distance $d(\vct{z}_{i}, \vct{z}_{j})$. Then, based on Definition~\ref{def:otd}, we formulate the OTD of feature representation $\vct{z}_{i}$ over $\mathcal{Q}_{i}$ as follows:
\begin{align}
\mat{Q}_{i,j} = \frac{\exp\left(- d(\vct{z}_{i}, \vct{z}_{j}) \right)}{\sum\limits_{\substack{n=1\\n\neq i}}^N \exp\big(- d(\vct{z}_{i}, \vct{z}_{n}) \big)}, \ j \in \{1, \ldots, N\},
\label{eq:q_Z}
\end{align}
where $\mat{Q}_{i, :} \in \mathbb{R}^{1 \times N}$, and $\mat{Q}_{i, j}$ is the probability that $\mat{z}_{i}$ picks $\mat{z}_{j}$ as its neighbor in latent space. Hence, the feature representations and corresponding ordered labels are embedded in a space of the same dimension, \ie, $\mat{Q}_{i, :} \in \mathbb{R}^{1 \times N}$ and $\mat{P}_{i, :} \in \mathbb{R}^{1 \times N}$. Then, we can let $\mat{Q}_{i, :}$ approximate $\mat{P}_{i, :}$ to learn the ordianl relations.

In practice, calculating the $N$-dimensional OTDs for all the samples can be computationally burdensome when the number of samples is too large. Therefore, we focus on the optimization problem with respect to a mini-batch of samples, which is consistent with mini-batch-based stochastic gradient descent. Specifically, we have $\mat{Q}_{i, :} \in \mathbb{R}^{1 \times N_{B}}$ and $\mat{P}_{i, :} \in \mathbb{R}^{1 \times N_{B}}$, \ie, $\mat{Q} \in \mathbb{R}^{N_{B} \times N_{B}}$, $\mat{P} \in \mathbb{R}^{N_{B} \times N_{B}}$, where $N_{B}$ is the mini-batch size. 
Then, ordinal relation learning for feature representations can be conducted by measuring the KL-divergence between $\mat{P}$ and $\mat{Q}$:
\begin{align}
\mathcal{L}_\mathrm{KL} \triangleq \frac{1}{N_{B}} \sum_{i = 1}^{N_{B}} \KL{\mat{P}_{i, :}}{\mat{Q}_{i, :}}.
\label{eq:kl}
\end{align}

\subsection{Prototype-Constrained Convex
Programming}
\subsubsection{Ordinal prototype constraint} During training, more than one sample will belong to the same rank, and then the label set can be regarded as a set of non-overlapping classes.
Here, we refer to $N_{K}$ as the number of classes within one mini-batch, and $k$ is the index of the $k$-th class.
Instead of tackling samples of the same classes independently within a mini-batch, we propose to calculate OTD for $y_{k}$, yielding the ordinal prototype constraint: $\mat{P}_{k,:} = \frac{1}{|\mathcal{U}(k)|} \sum_{u \in \mathcal{U}(k)} \mat{Q}_{u,:}$, where $\mathcal{U}(k)$ denotes a set of all the samples belonging to the $k$-th class, and $N_{B} = \sum_{k=1}^{N_{K}} |\mathcal{U} (k)|$. Due to the ordinal information preserved by $\mat{P}_{k,:}$, this constraint certainly maintains ordinal information for $\frac{1}{|\mathcal{U}(k)|} \sum_{u \in \mathcal{U}(k)} \mat{Q}_{u,:}$. Therefore, we obtain the following optimization problem:

\begin{align}
\min_{\mat{\theta}}&  \quad \frac{1}{N_{K}} \sum_{k = 1}^{N_{K}} \frac{1}{|\mathcal{U}(k)|} \sum_{u \in \mathcal{U}(k)} \KL{\mat{P}_{k,:}}{\mat{Q}_{u, :}}  \notag\\
\mathrm{s.t.}&\quad  \mat{P}_{k, :} = \frac{1}{|\mathcal{U}(k)|} \sum_{u \in \mathcal{U}(k)} \mat{Q}_{u,:}.
\label{eq:kl_constraint}
\end{align}

Since $\mat{P}_{k,:}$ inherits the ordinal constraint in Eq.~\eqref{eq:or_constraint} so that it is a \emph{toset} as proven in Proposition~\ref{proposition:toset}, the model will learn to cluster feature representations within a mini-batch. In other words, the mean OTD of $\mat{Q}_{u,:}$, where $u\in \mathcal{U}(k)$, can approach that of the prototypical OTD $\mat{P}_{k,:}$; interestingly, this results in more compact manifolds
Consequently, minimizing the constrained optimization problem in Eq.~\eqref{eq:kl_constraint} guarantees consistency between feature representations and the corresponding ordinal labels in latent space.

\subsubsection{Convex
programming with dual decomposition}
Although $\mat{P}$ and $\mat{Q}$ have the same dimensions, they are derived from different domains, \ie, ordinal labels and images, respectively; optimizing $\KL{\mat{P}}{\mat{Q}}$ directly is challenging due to the internal covariate shift. Therefore, we propose using strong duality to decompose the optimization in Eq.~\eqref{eq:kl_constraint}. Specifically, we infer probability distributions $\widetilde{\mat{P}}$ and $\widetilde{\mat{Q}}$ that are close to $\mat{P}$ and $\mat{Q}$, respectively. Then, the optimization problem is converted as follows:
\begin{align}
\min_{\widetilde{\mat{P}}, \widetilde{\mat{Q}}} &\quad \frac{1}{N_{K}}\sum_{k=1}^{N_{K}} \Big[ \KL{\widetilde{\mat{P}}_{k, :}}{\mat{P}_{k, :}} \!+\! \frac{1}{|\mathcal{U}(k)|}\sum_{u=1}^{|\mathcal{U}(k)|} \KL{\widetilde{\mat{Q}}_{u, :}}{\mat{Q}_{u, :}} \Big] \notag \\
  \mathrm{s.t.}&\quad \widetilde{\mat{P}}_{k, :} = \frac{1}{|\mathcal{U}(k)|} \sum_{u \in \mathcal{U}(k)} \widetilde{\mat{Q}}_{u, :},
\label{eq:dual_kl}
\end{align}
where $\widetilde{\mat{P}}_{k, :}$ and $\widetilde{\mat{Q}}_{u, :}$ are probability simplexes corresponding to class $y_{k}$ and the $u$-th sample, respectively. To optimize the above convex program, we assign each prototype constraint $\widetilde{\mat{P}}_{k, :} = \frac{1}{|\mathcal{U}(k)|} \sum_{u \in \mathcal{U}(k)} \widetilde{\mat{Q}}_{u, :}$ with a Lagrange multiplier $\lambda_{k}$. Note that the total number of $\lambda_{k}$ is equal to $C$, the number of classes in a whole dataset. Finally, we need to optimize the following objective:
\begin{align}
\min_{\widetilde{\mat{P}}, \widetilde{\mat{Q}}}   \frac{1}{N_{K}}&\sum_{k=1}^{N_{K}} \Big[ \KL{\widetilde{\mat{P}}_{k, :}}{\mat{P}_{k, :}} + \frac{1}{|\mathcal{U}(k)|}\sum_{u=1}^{|\mathcal{U}(k)|} \KL{\widetilde{\mat{Q}}_{u, :}}{\mat{Q}_{u, :}} \Big] \notag \\
 +& \sum_{k=1}^{N_{K}} \lambda_{k} \Big[ \underbrace{ \widetilde{\mat{P}}_{k, :} - \frac{1}{|\mathcal{U}(k)|} \sum_{u \in \mathcal{U}(k)} \widetilde{\mat{Q}}_{u, :}}_\text{Prototype \ constraint} \Big].
\label{eq:dual_kl_lambda}
\end{align}

Following~\cite{ron2022dual}, we can obtain the optimal solutions for Eq.~\eqref{eq:dual_kl}:

\begin{proposition}
The optimal solutions $\widetilde{\mat{P}}_{k, :}$ and $\widetilde{\mat{Q}}_{u, :}$ for convex programming in Eq.~\eqref{eq:dual_kl} are derived as: 
\begin{align}
&\widetilde{\mat{P}}_{k, j} = \frac{\mat{P}_{k, j} e^{- \lambda_{k}}}{\sum_{s} \mat{P}_{k, s} e^{- \lambda_{s}}}, \\
&\widetilde{\mat{Q}}_{u, j} = \frac{\mat{Q}_{u, j} e^{\lambda_{k}}}{\sum_{t}\sum_{s \in \mathcal{U}(t)} \mat{Q}_{u, s} e^{\lambda_{t}}}.
\end{align}

Since $\widetilde{\mat{P}}_{k, :}$ and $\widetilde{\mat{Q}}_{u, :}$ are reparameterized via dual variable $\lambda$, they maintain the consistent ordinal-prototype constraint: $\widetilde{\mat{P}}_{k, :} = \frac{1}{|\mathcal{U}(k)|} \sum_{u \in \mathcal{U}(k)} \widetilde{\mat{Q}}_{u, :}$.
\label{proposition:reparam}
\end{proposition}

Proposition~\ref{proposition:reparam} indicates that we only need to learn the dual variables $\lambda_{k}$; then, the prototypical constraint can be maintained. Hence, $\lambda_{k}$ serves as the dual witness to the optimality of $\widetilde{\mat{P}}_{k, :}$ and $\widetilde{\mat{Q}}_{u, :}$. Then, we can optimize KL-divergence between the reparameterized $\widetilde{\mat{P}}_{k, :}$ and $\widetilde{\mat{Q}}_{u, :}$:
\begin{align}
\mathcal{L}_\mathrm{KL}^\mathrm{Dual} \triangleq \frac{1}{N_{K}} \sum_{k = 1}^{N_{K}} \Big[ \frac{1}{|\mathcal{U}(k)|} \sum_{u\in \mathcal{U}(k)} \KL{\widetilde{\mat{P}}_{k, :}}{\widetilde{\mat{Q}}_{u, :}} \Big].
\end{align}

In addition, we find that the values of $\lambda_{k}$ of different classes are identical during training, which makes it difficult to identify the learning state of inter-class prototypes in Eq.~\eqref{eq:dual_kl_lambda}. To address this issue, we minimize an entropy regularization term to enforce the learning of informative $\lambda_{k}$ values, defined as:
\begin{align}
\mathcal{L}_{\text{Ent}} = - \sum_{k=1}^{N_{K}} \lambda_k \log (\lambda_k).
\label{eq:entropy}
\end{align}

Note that the entropy $\mathcal{L}_{\text{Ent}}$ is a commonly used regularization, and per the definition of Lagrange duality, the dual variable $\lambda$ is positive. As $\lambda_{k}$ is learned during training and reflects the prototype alignment difficulty for the $k$-th class shown in Eq.~\eqref{eq:dual_kl_lambda}, minimizing the entropy term across all classes regularizes the model to have varying abilities for ensuring class-specific prototype constraints.

\begin{algorithm}[t]
\caption{Training algorithm of \methodname}
\label{alg:algorithm}
\textbf{Input}: Training data $\mathcal{X}$, labels $\mathcal{Y}$, mini-batch size: $N_{\text{B}}$, number of mini-batches: $N_{\text{Mini-batch}}$, CNN feature extractor $f_{\mat{\theta}}$, Linear layer $g_{\mat{\phi}}$, total epochs: $T$, hyper-parameters $\alpha$, $\beta$. \\
\textbf{Parameter}: Parameters of CNN and final Linear layer: $\mat{\theta}$ and $\mat{\phi}$, dual variables: $\lambda_{k}$ where $k$ is class index. \\
\textbf{Output}: optimal $\mat{\theta}^{*}$, $\mat{\phi}^{*}$, and $\lambda_{k}^{*}$.
\begin{algorithmic}[1] 
\STATE Initialize $\mat{\theta}$ and $\mat{\phi}$ as weights from pre-trained models such as VGG-16, $\lambda_{k}$ as $1$.
\FOR{epoch $\leq$ $T$}
    
    \FOR{$b \leq N_{\text{Mini-batch}}$}
        \STATE Sampling a mini-batch  $\{ (\mat{x}_{i}, y_{i}) \}_{i=1}^{N_{B}}$
        \FOR{$i \leq N_{\text{B}}$}
            \STATE $\vct{z}_{i} = f_{\mat{\theta}} (\mat{x}_{i})$
            \ENDFOR
        \STATE Compute OTDs  $\mat{Q}_{i,:}$ and $\mat{P}_{i,:}$ via Eqs.~\eqref{eq:p_Y} and~\eqref{eq:q_Z}.
        \STATE 
        $ \widetilde{\mat{P}}_{k, :} = \frac{\mat{P}_{k, j} e^{- \lambda_{k}}}{\sum_{s} \mat{P}_{k, s}  e^{- \lambda_{s}}}$, \\ 
        $\widetilde{\mat{Q}}_{u, :} = \frac{\mat{Q}_{u, j}  e^{\lambda_{k}}}{\sum_{t}\sum_{r\in\mathcal{U}(t)} \mat{Q}_{j,r}  e^{\lambda_{t}}}$.
        \STATE Calculate the objective function $\mathcal{L}$ in Eq.~\eqref{eq:loss} 
        \STATE Gradient back-propagation: $\frac{\partial{\mathcal{L}}}{\partial{\mat{\theta}}}$, $\frac{\partial{\mathcal{L}}}{\partial{\mat{\phi}}}$, $\frac{\partial{\mathcal{L}}}{\partial{\lambda}}$
    \ENDFOR
\ENDFOR
\end{algorithmic}
\end{algorithm}

Consequently, the final learning objective is:
\begin{align}
\mathcal{L} = \mathcal{L}_\text{OR} + 
 \alpha \mathcal{L}_\mathrm{KL}^\mathrm{Dual} + \beta \mathcal{L}_{\text{Ent}},
\label{eq:loss}
\end{align}
where $\mathcal{L}_\text{OR}$ denotes the losses of existing OR methods that can be combined with \methodname, and $\alpha$ and $\beta$ are hyper-parameters that adjust the magnitude of the loss. The detailed training procedure is shown in Algorithm~\ref{alg:algorithm}.

\noindent\emph{Relationship Between $N_{B}$ and $N_{K}$}:\quad
Recall that $N_{B}$ is the mini-batch size and $N_{K}$ is the number of classes involved in the mini-batch. Generally, when the model is trained with stochastic gradient descent, the following conditions hold for the randomly sampled mini-batches:
\begin{itemize}
  \item $N_{K} < N_{B}$. This condition indicates that there exists at least one class containing more than one sample; hence, for these classes, the prototype alignment in Eq.~\eqref{eq:dual_kl_lambda} works.
  \item $N_{K} = N_{B}$. This condition indicates that all the $N_{B}$ samples have distinct labels, and the prototype constraint will be unable to induce prototype alignment, as will be discussed further in the experiments.
\end{itemize}


\section{Experiments}
\label{sec:experiment}
In this section, we evaluate our method on four different applications: facial age estimation, medical image classification, historical image dating, and image aesthetic assessment tasks.

\subsection{Datasets for Four Different Applications}
\subsubsection{Facial age estimation} The 
\textbf{MORPH}~\cite{morph} dataset contains 55,134 human facial images, each of which is labeled by real age of the individual, rating from 16 to 77 years. Following~\cite{DRFs,CORFs}, we conduct 5 runs with random splits in the dataset such that 80\% is used for training and 20\% for testing, and we report the average results. \textbf{FG-Net}~\cite{fgnet} includes 1,002 facial images of 82 persons, each of whom is represented by more than 10 ages. The images vary with respect to pose, lighting, and facial expression. \textbf{Adience}~\cite{6906255} contains 26,580 facial images of 2,284 persons, and the ordinal labels correspond to eight age ranges: 0-2, 4-6, 8-13, 15-20, 25-32, 38-43, 48-53, and $>$60 years.

\subsubsection{Medical disease progression prediction} \textbf{LIDC-IDRI} is a computed tomography (CT) dataset for lung nodule classification consisting of the CT scans of 1,010 patients. Each nodule is rated from 1 to 5, representing the malignancy progression. Following~\cite{UDM,lei2022meta}, we label nodules with average scores lower than $2.5$ as benign, $2.5\sim3.5$ as unsure, and higher than $3.5$ as malignant. The \textbf{BUSI}~\cite{busi} dataset is used for ultrasound (US)-based breast cancer classification and segmentation; it includes 780 images comprising three categories: 133 normal, 487 benign, and 210 malignant cases. The diabetic retinopathy (DR) dataset includes high-resolution fundus images of 17,563 patients, each of whom provided left and right fundus images tagged with one of five ordinal labels: no DR, mild DR, moderate DR, severe DR, and proliferative DR. Following~\cite{NSB}, we conduct 10-fold cross-validation with these data.

\begin{table*}[b]
\caption{Experimental results on facial age estimation. Note that $l=5$ used for CS.}
\centering
\begin{tabular*}{1\linewidth}{@{\extracolsep{\fill}}lllllllll}
\shline
\multirow{2}{*}{Methods} & \multicolumn{2}{c}{MORPH} && \multicolumn{2}{c}{FG-NET} && \multicolumn{2}{c}{Adience} \\ \cline{2-3}  \cline{5-6}  \cline{8-9}
& MAE$\downarrow$ & CS(\%)$\uparrow$ && MAE$\downarrow$ & CS(\%)$\uparrow$ && MAE$\downarrow$ & Acc.(\%)$\uparrow$ \\
\midrule
MV~\cite{Mean-variance}              & 2.41 & 90.0  && 4.10 & 78.2 && 0.48 & 66.9 \\
DRFs~\cite{DRFs}                     & 2.19 & 91.3  && 3.85 & 80.6 && 0.45 & 60.3 \\
Poisson~\cite{Unimodal}              & 2.66 & 88.1  && 3.47 & 77.6 && 0.46 & 59.9 \\
SORD~\cite{SORD}                     & 2.39 & 89.4  && 2.42 & 82.9 && 0.49 & 59.6 \\
POE~\cite{POE}                       & 2.35 & 92.4   && 2.41 & 81.8 && 0.47 & 60.5 \\
CORFs~\cite{CORFs}                   & 2.17 & 93.0  && 2.68 & 86.8 && 0.44 & 72.2 \\
Beta~\cite{Beta}                     & 2.33 & 91.5   && 2.53  & 79.5 && 0.45 & 76.4 \\
\hline
\textbf{\methodname} + MV  & 2.13\gbf{$\downarrow$ 0.28}   & \bf{95.6}\gbf{$\uparrow$ 5.6}      && 2.13\gbf{$\downarrow$ 1.96} & 92.8\gbf{$\uparrow$ 14.6} && \bf{0.39}\gbf{$\downarrow$ 0.09} & \bf{83.6}\gbf{$\uparrow$ 16.7} \\
\textbf{\methodname} + Poisson           & 2.31\gbf{$\downarrow$ 0.35}        & 94.1\gbf{$\uparrow$ 6.0}      && 2.07\gbf{$\downarrow$ 1.40} & 94.3\gbf{$\uparrow$ 16.7} && 0.44\gbf{$\downarrow$ 0.02} & 75.6\gbf{$\uparrow$ 15.7} \\
\textbf{\methodname} + SORD              & \bf{2.10}\gbf{$\downarrow$ 0.29}        & \bf{95.6}\gbf{$\uparrow$ 6.2} && \bf{2.02}\gbf{$\downarrow$ 0.40}   & \bf{95.1}\gbf{$\uparrow$ 12.2} && 0.41\gbf{$\downarrow$ 0.07} & 76.1\gbf{$\uparrow$ 16.5}\\
\textbf{\methodname} + POE               & 2.14\gbf{$\downarrow$ 0.21}        & 94.9\gbf{$\uparrow$ 2.5}      && 2.15\gbf{$\downarrow$ 0.26} & 90.7\gbf{$\uparrow$ 7.9} && 0.42\gbf{$\downarrow$ 0.05} & 75.4\gbf{$\uparrow$ 14.9} \\
\shline
\end{tabular*}
\label{tab:age}
\end{table*}
\subsubsection{Historical image dating (\emph{HID})} The \textbf{HID}~\cite{HID} is used to estimate ``ages'' of color images that were taken from five different decades, \ie, the 1930s to 1970s, each of which is regarded as one class in this benchmark. Following the settings used in~\cite{POE,CNNPOR}, we conduct 10-fold cross-validation with these data and report the mean results.

\subsubsection{Image aesthetics assessment} The aesthetics with attributes database (\textbf{AADB}) contains 10,000 natural images that are labeled by average aesthetic scores ranging from $0$ to $1$~\cite{aadb}. We conduct binary classification to predict whether the scores are greater or less than $0.5$. Following~\cite{CORFs}, we use 9,000 images for training and 1,000 for testing.

\subsection{Implementation Details}
Following~\cite{SORD, CORFs}, in our experiments, we use the VGG-16~\cite{vgg} network pretrained on ImageNet as the backbone to ensure fair comparisons. For POE and \methodname+ POE, the backbone network is VGG19~\cite{POE}. The feature representation is obtained by \{Linear-ReLU-Dropout\}$\times2$, as shown in Fig.~\ref{fig:framework}, and its dimension is $4,096$. 
The training mini-batch size is set to $32$. 
All the training images for facial age estimation were first resized to the size $256 \times 256$ and then randomly cropped to $224 \times 224$. Augmentation approaches, including random rotation and flipping were also utilized during training. The total number of epochs is $50$, and the learning rate is initially set to $0.0001$ and decayed by $0.1$ every $20$ epochs. We use Adam optimizer with a weight decay of  $0.0001$~\cite{adam}. $\alpha$ and $\beta$ are empirically set to $10$ and $1$, respectively. All experiments are implemented using the PyTorch~\cite{pytorch} framework and trained with one NVIDIA A100 GPU.

\subsection{Evaluation Metrics}
In our experiments, we report the evaluation metrics according to previous works that use specific metrics for different tasks. We calculate the mean absolute error (MAE) for all the experiments other than the image aesthetics assessment. In addition, we compare the cumulative score (CS) for the age estimations, which contains a hyperparameter $l$ indicating that a result with an MAE less than $l$ is regarded as the correct one.
We adopt $l = 5$, the same as~\cite{DRFs,CORFs}. For the medical disease prediction datasets, the Adience dataset and the historical image dating dataset, we report the classification accuracy. Following~\cite{CORFs,niu2016ordinal}, we report the Pearson linear correlation coefficient (PLCC) and Spearman’s rank correlation coefficient (SRCC) for the AADB dataset.

\begin{table*}[ht]
\caption{Experimental results on medical disease progression prediction in terms of MAE ($\times 10^{-2}$) and Accuracy.}
\centering
\begin{tabular*}{1\linewidth}{@{\extracolsep{\fill}}llllllllll}
\shline
\multirow{2}{*}{Methods} & \multicolumn{2}{c}{LIDC-IDRI} && \multicolumn{2}{c}{BUSI} && \multicolumn{2}{c}{DR} \\ \cline{2-3}  \cline{5-6}  \cline{8-9}
& MAE$\downarrow$ & Acc.(\%)$\uparrow$ && MAE$\downarrow$  & Acc.(\%)$\uparrow$ && MAE$\downarrow$  & Acc.(\%)$\uparrow$ \\
\midrule
Poisson~\cite{Unimodal}              & 52.29 & 54.2 && 23.57       &75.2   &&   37.43& 79.6   \\
MV~\cite{Mean-variance}              & 49.26 & 54.8 && 21.03       &77.6   &&   27.63& 81.9   \\
DRFs~\cite{DRFs}                     & 49.43 & 54.2 && 20.70       &75.6   &&   29.45& 80.5   \\
NSB~\cite{NSB}                       & 51.46 & 55.3 && 21.66       &76.4   &&   32.00& 84.2   \\
SORD~\cite{SORD}                     & 50.38 & 54.0 && 20.36       &78.2   &&   26.81& 82.1   \\
UDM~\cite{UDM}                       & 51.84 & 54.8 && 24.84       &77.7   &&   26.38& 81.2   \\
POE~\cite{POE}                       & 52.48 & 54.6 && 21.16       &75.0   &&   29.25& 80.9   \\
\hline
\textbf{\methodname}  + MV & \textbf{48.82}\gbf{$\downarrow$ 0.44} & 55.5\gbf{$\uparrow$ 0.7}   &&   20.21\gbf{$\downarrow$ 0.82} &76.3\rbf{$\downarrow$ 1.3}   &&   26.61\gbf{$\downarrow$ 1.02} & 82.3\gbf{$\uparrow$ 1.1} \\
\textbf{\methodname}  + Poisson & 50.03\gbf{$\downarrow$ 2.26} & 54.6\gbf{$\uparrow$ 0.4}   &&   21.64\gbf{$\downarrow$ 1.93} & 79.6\gbf{$\uparrow$ 4.4}   &&   32.67\gbf{$\downarrow$ 4.76} & 79.5\rbf{$\downarrow$ 0.1} \\
\textbf{\methodname}  + SORD & 49.58\gbf{$\downarrow$ 0.80} & \textbf{55.8}\gbf{$\uparrow$ 1.8}    &&    \textbf{18.47}\gbf{$\downarrow$ 1.89} & \textbf{82.2}\gbf{$\uparrow$ 4.0}   &&   \textbf{25.04}\gbf{$\downarrow$ 1.37} & 83.3\gbf{$\uparrow$ 1.2} \\
\textbf{\methodname}  + POE & 51.43\gbf{$\downarrow$ 1.05} & 54.8\gbf{$\uparrow$ 0.2}   &&   20.38\gbf{$\downarrow$ 0.78} & 80.3\gbf{$\uparrow$ 5.3}   &&   27.18\gbf{$\downarrow$ 2.07} & \textbf{84.3}\gbf{$\uparrow$ 3.4}  \\
\shline
\end{tabular*}
\label{tab:medical}
\end{table*}

\begin{table}[ht]
\caption{Experimental results on HID dataset.}
\centering
\begin{tabular*}{1\linewidth}{@{\extracolsep{\fill}}lcc}
\shline
Methods & MAE$\downarrow$ & Acc.(\%)$\uparrow$ \\
\midrule
Palermo~\cite{HID} & 0.93$\pm$0.08 & 44.92$\pm$3.69 \\
CNNPOR~\cite{CNNPOR} & 0.82$\pm$0.05 & 50.12$\pm$2.65 \\                   
GP-DNNOR~\cite{GP-DNNOR} & 0.76$\pm$0.05 & 46.60$\pm$2.98 \\ 
MV~\cite{Mean-variance} & 0.72$\pm$0.05 & 53.73$\pm$2.91 \\
Poisson~\cite{Unimodal} & 0.83$\pm$0.06 & 51.38$\pm$3.42 \\
SORD~\cite{SORD} & 0.70$\pm$0.03 & 54.36$\pm$3.11 \\
POE~\cite{POE} & 0.67$\pm$0.04 & 54.68$\pm$3.21 \\
\hline
\textbf{\methodname}  + MV & 0.65$\pm$0.03 & 54.46$\pm$3.12 \\
\textbf{\methodname}  + Poisson & 0.77$\pm$0.04 & 53.94$\pm$3.32 \\
\textbf{\methodname}  + SORD & \textbf{0.60$\pm$0.03} & 55.69$\pm$2.86 \\
\textbf{\methodname}  + POE & 0.63$\pm$0.01 & \textbf{55.83$\pm$3.07} \\
\shline
\end{tabular*}
\label{tab:hid}
\end{table}

\subsection{Main Results}
For the quantitative comparisons, we conducted experimental comparisons between a number of existing, state-of-the-art ordinal regression methods and their counterparts combined with \methodname. We mainly adopted four baseline methods: Mean-Variance (MV)~\cite{Mean-variance}, Poisson~\cite{Unimodal}, SORD~\cite{SORD}, and POE~\cite{POE}. In Table~\ref{tab:age}, the proposed \methodname consistently improves the ordinal estimation performance of these methods in terms of MAE, CS, and accuracy. 
MV computes the mean and variance loss of the output vector to control the uncertainties in the output space. \methodname + MV achieved the best MAE values on the MORPH and Adience datasets, which confirms that \methodname enables alignment of the learned mean and variance to the ordinal information in label space. For the FG-Net dataset, which contains differences in facial  expression and lighting, \methodname + SORD performs better than the other methods. This suggests that the better ordinal representation can fit more appropriately with the soft labels constructed by SORD. More importantly, \methodname improves the CS scores and accuracy by a large margin, indicating that the learned ordinal representations guarantee better results after the final linear mapping layer.

Medical disease progression is intrinsically discriminated by ordinal information that presents disease progression. Table~\ref{tab:medical} also shows the consistent improvements imparted by \methodname; however, the improved performance margins are not as substantial as those seen for age estimation. We argue that the lower number of ordinal labels  in medical datasets (< 10) than in the age datasets such as MORPH (62 classes) is not beneficial for modeling real manifold of the ordinal relations generated by the proposed OTD. Moreover, the ground-truth labels of the medical datasets are prone to bias due to differences among the experts making the labels and therefore may be incorrect, affecting the learning of ordinal relations.

\begin{table}[ht]
\caption{Experimental results on the AADB dataset.}
\centering
\begin{tabular*}{1\linewidth}{@{\extracolsep{\fill}}lll}
\shline
Methods & SRCC$\uparrow$ & PLCC$\uparrow$ \\
\midrule
MV~\cite{Mean-variance} & 0.4304 & 0.4467 \\
Poisson~\cite{Unimodal}  & 0.5032 & 0.5396 \\
SORD~\cite{SORD}  & 0.6635 & 0.6782 \\
POE~\cite{POE}  & 0.6208 & 0.6292 \\
OR-CNN~\cite{niu2016ordinal} & 0.4370 & 0.4388 \\
CORF~\cite{CORFs} & 0.6770 & 0.6829 \\
\hline
\textbf{\methodname}  + MV  & 0.5443\gbf{$\uparrow$ 0.1139} & 0.5632\gbf{$\uparrow$ 0.1165} \\
\textbf{\methodname}  + Poisson  & 0.5152\gbf{$\uparrow$ 0.0120} & 0.5273\rbf{$\downarrow$ 0.0159} \\
\textbf{\methodname}  + SORD  & \bf{0.6832}\gbf{$\uparrow$ 0.0197} & \bf{0.6993}\gbf{$\uparrow$ 0.0211} \\
\textbf{\methodname}  + POE  & 0.6413\gbf{$\uparrow$ 0.0205} & 0.6508\gbf{$\uparrow$ 0.0216} \\
\shline
\end{tabular*}
\label{tab:aadb}
\end{table}

For historical image dating, following~\cite{POE}, we report the means and standard deviations in Table~\ref{tab:hid}. A similar conclusion can be drawn as above: the existing ordinal regression methods consistently demonstrate enhanced ordinal relation learning when combined with \methodname. HID differs from the previous two tasks in that the contents of the images of different classes/decades have no fixed objects that underlie the  ordinal information; the discriminative information among different decades are tones,building styles, \etc. Although POE infers  uncertainties through the learning of the means and covariances of multi-variable Gaussian of feature representations, the ordinal constraint in Eq.~\eqref{eq:or_constraint} is only preserved among hard examples in a mini-batch~\cite{POE}. In contrast, the OTDs of \methodname maintain the  ordinal information in the embedding space for all the samples. Most importantly, \methodname imposes the ordinal relations residing in the labels directly on a neighborhood of feature representations, \ie, a mini-batch of samples. Therefore, the ordinal embeddings of \methodname can alleviate the influence of decision bias on certain objects while capturing the critical features of different decades.

For image aesthetic estimation on the AADB, Table~\ref{tab:aadb} demonstrates the superiority of SORD, POE, and their \methodname counterparts over MV and Poisson. This indicates the importance of modeling ordinal relations with respect to the latent space and label space. It can be seen that the improvements in performance with the AADB are also not as substantial as those in the facial age estimation, which we conclude is due to the bias in the scores of the AADB made by different humans; therefore, the ordinal relations may be incorrect.

\begin{figure*}[ht]
\centering
\includegraphics[width=1.\linewidth]{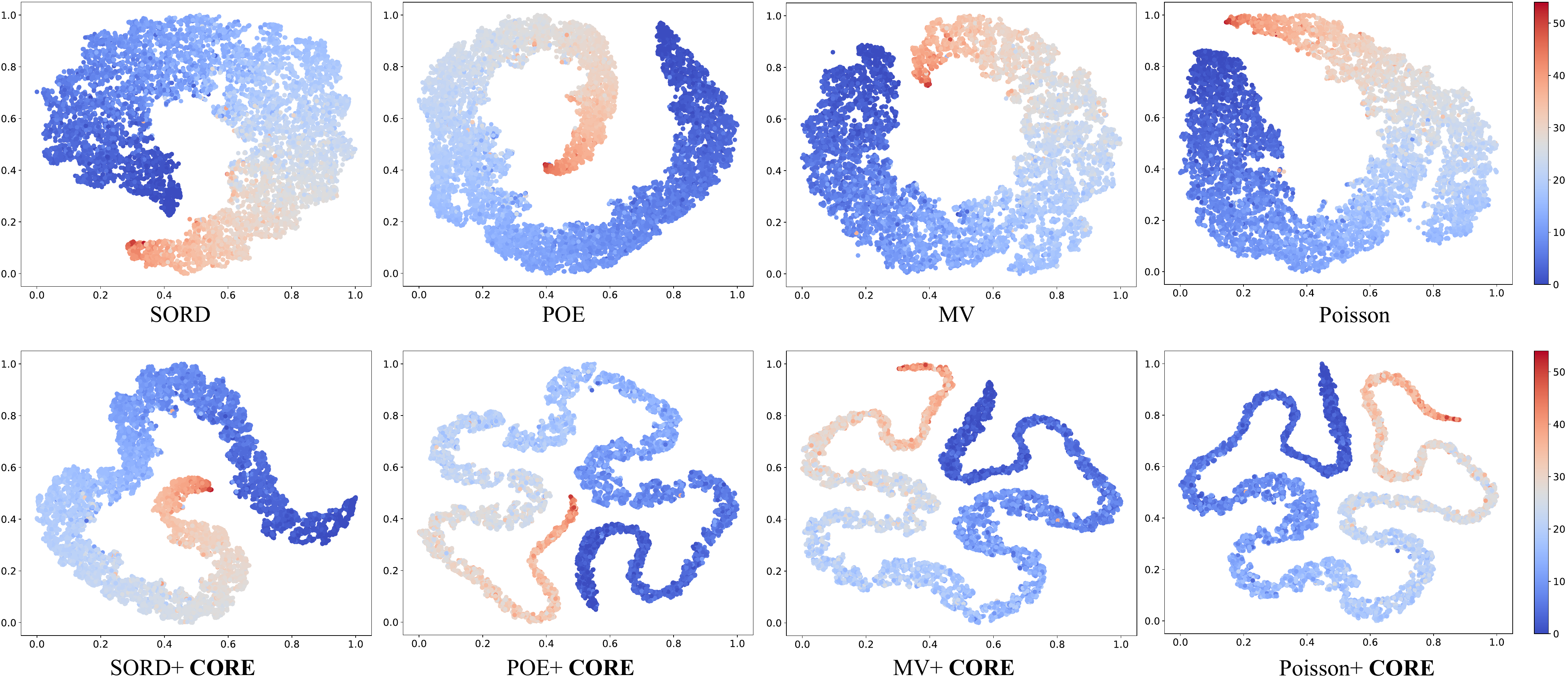}
\caption{Comparisons of t-SNE results obtained by different methods on MORPH test set. The values on the color bar represent the ordinal labels. }
\label{fig:visualizations}
\end{figure*}
\begin{figure*}[ht]
\centering 
\includegraphics[width=1.\linewidth]{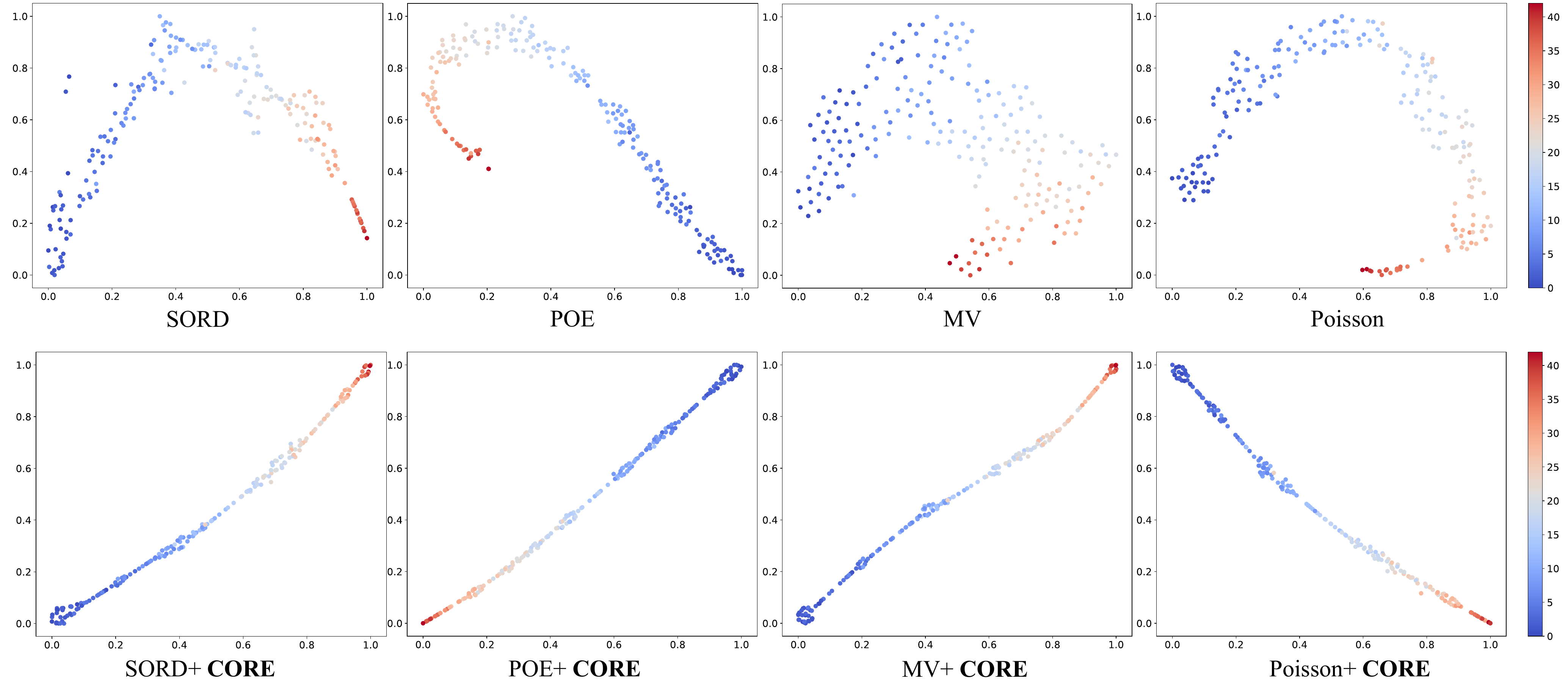}
\caption{Comparisons of t-SNE results obtained by different methods on randomly sampled 200 samples in MORPH test set. The values on the color bar represent the ordinal labels. }
\label{fig:local_visualizations}
\end{figure*}
\subsection{Visualization of Consistent Ordinal Representation}
To further highlight the superiority of \methodname,  we visualize the feature representations of the four baseline methods with and without \methodname using t-SNE~\cite{tsne} in Fig.~\ref{fig:visualizations}. All four methods show particular ordinal structures in the feature space. Although their learning strategies can encourage ordinal relation learning in latent space, they lack intraclass compactness.
With our \methodname, all the results can be adjusted to be more compact and exhibit a nearly linear manifold among neighbor orders.

Furthermore, \methodname preserves not only the ordinal relationship globally but also maintains a nearly one-dimensional manifold for local observations. In Fig.~\ref{fig:local_visualizations}, we provide the visualizations of randomly selected local samples from the MORPH test set. We can see that \methodname enables more compactness in the latent space than baseline methods. Although the results of the  Mean-Variance and Poisson methods are relatively discrete, \methodname also preserves more continuous manifold structures. More t-SNE results on other datasets are provided in the supplementary material.

\begin{figure*}[ht]
\centerline{\includegraphics[width=1.0\linewidth]{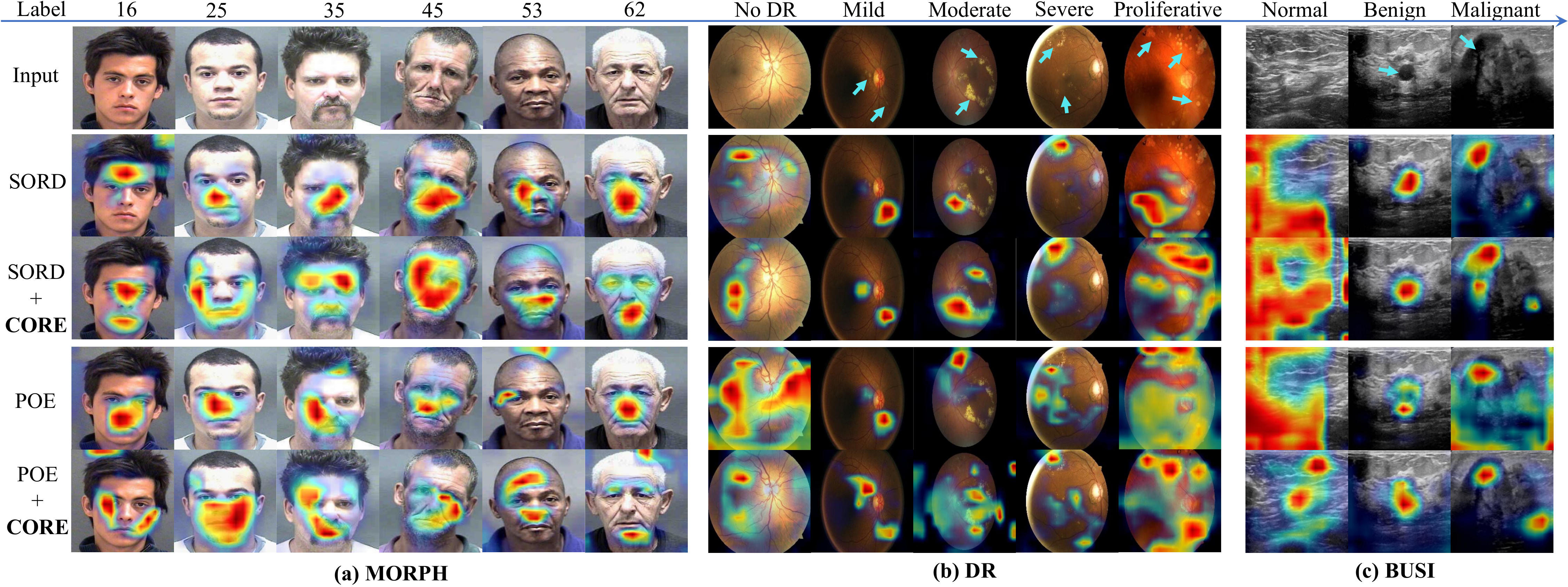}}
\caption{Grad-CAMs on MORPH, DR, and BUSI datasets. We have manually highlighted the lesions of DR and BUSI with cyan arrows on input images. From left to right, the images are arranged in ascending order of labels.}
\label{fig:cams}
\end{figure*}

\begin{figure*}[ht]
\centerline{\includegraphics[width=1\linewidth]{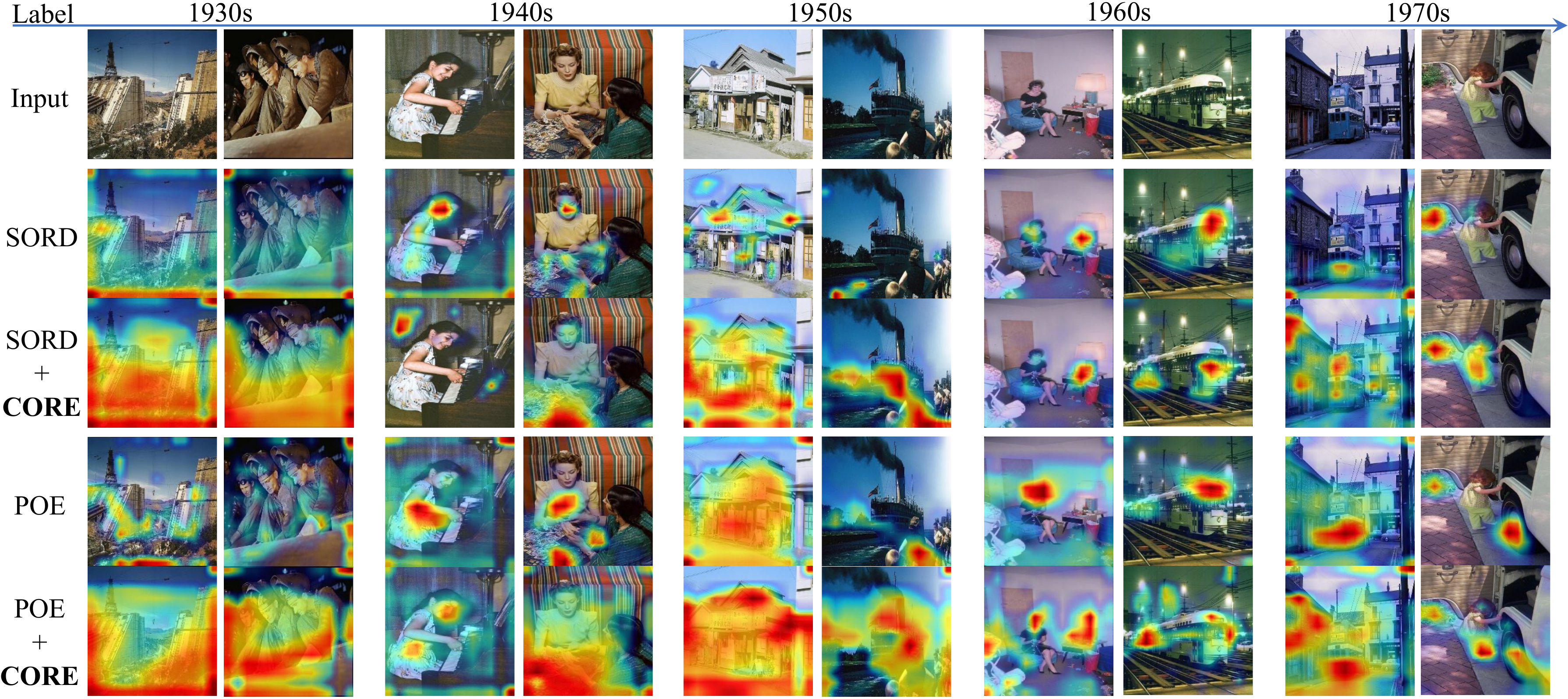}}
\caption{Grad-CAMs on HID dataset. For each decade, we selected two example images. From left to right, the images are arranged in ascending order of labels.}
\label{fig:cams_hid}
\end{figure*}

\subsection{Visual Interpretation of Ordinal Information}
To better understand the ordinal information learned by \methodname, we apply CNN interpretation methods to further investigate which kind of visual features are related to the real orders. We conducted both qualitative and quantitative experiments to compare different methods. 

\subsubsection{Qualitative results} We selected SORD and POE as baseline models and show the results of Grad-CAM~\cite{grad-cam} on the MORPH, DR, and BUSI datasets with increasing ordinal labels in Fig.~\ref{fig:cams}. We can see that the combination of \methodname and the baseline methods captures more accurate and more complete features such as wrinkles, beards, and hairs on facial images, diabetic lesions on fundus images, and tumors on ultrasound images. Further considering the t-SNE visualizations in Figs.~\ref{fig:visualizations} and~\ref{fig:local_visualizations}, we conclude that the proposed \methodname indeed aligns the features and corresponding ordinal labels. In Fig.~\ref{fig:cams_hid}, we can see that the visual attention across decades is placed on certain objects such as buildings, humans and trains, \etc, but there is no evidence that a particular kind of object dominates the differences across decades. From the perspective of humans, there is notable variation for different styles of either the whole image or some local areas. Nevertheless, \methodname consistently helps localize objects of interest that reflect the style of the image as completely as possible. Interestingly, we also find that the attention seems to be placed in random locations for the ``No DR'' class of DR dataset images and the ``Normal'' class of BUSI dataset images. We conjecture that because there are no explicit lesions or tumors in these two classes, it is difficult for the models to capture meaningful attention regions, a weakness that cannot be solved by advanced vision interpretation techniques such as Score-CAM~\cite{score-cam} and Group-CAM~\cite{group-cam}.

\subsubsection{Quantitative results} Following RISE~\cite{rise}, Score-CAM~\cite{score-cam}, and Group-CAM~\cite{group-cam}, we conduct \textbf{Insertion} and \textbf{Deletion} experiments on the four applications. Insertion gradually introduces class-related regions ($3.6\%$ pixels) of an original image to a blurred image according to the values of the saliency map. This process is repeated until the blurred image is fully recovered. In contrast, Deletion aims to replace related pixels ($3.6\%$) in a blurred image with those of the corresponding original image. We report the AUC values of the classification score in Table~\ref{tab:ins_del}. For the four applications, \methodname generates more discriminative regions that are related to model decision-making. We observe that the Insertion values for the DR and BUSI datasets obtained with and without \methodname are similar, which is attributed to \textit{(i)} a relatively larger number of samples and \textit{(ii)} randomly generated attention maps for healthy classes, \ie, ``No DR'' and ``Normal'', which trigger ambiguous decisions when gradually introducing attention regions.

\begin{table*}
\caption{Comparative evaluation in terms of deletion  and insertion  AUC on the testing sets of MORPH, DR, BUSI, HID, and AADB.}
\centering
\begin{tabular*}{1\linewidth}{@{\extracolsep{\fill}}cccccccccccccccc}
\shline
\multirow{2}{*}{AUC} && \multicolumn{2}{c}{MORPH} && \multicolumn{2}{c}{DR} && \multicolumn{2}{c}{BUSI} && \multicolumn{2}{c}{HID} && \multicolumn{2}{c}{AADB} \\ 
\cline{3-4}  \cline{6-7} \cline{9-10} \cline{12-13} \cline{15-16}
&& SORD~\cite{SORD} & +\methodname && SORD & +\methodname && SORD~\cite{SORD} & +\methodname && SORD~\cite{SORD} & +\methodname && SORD~\cite{SORD} & +\methodname \\

\midrule
\textbf{Insertion} ($\uparrow$)   && 52.6 & \bf{55.3} && 58.7 & \bf{58.8} && \bf{44.7} & 43.4      && 8.10 & \bf{16.5} && 60.7 & \bf{62.5} \\
\textbf{Deletion} ($\downarrow$)  && 40.1 & \bf{38.9} && 57.5 & \bf{45.1} && 43.4      & \bf{36.9} && 6.19 & \bf{12.7} && 59.4 & \bf{43.2} \\
\textbf{Over-all} ($\uparrow$)    && 12.5 & \bf{16.4} && 1.2  & \bf{13.7} && 1.3       & \bf{6.5}  && 1.91 & \bf{3.8} && 1.3 & \bf{19.3} \\
\shline
\end{tabular*}
\label{tab:ins_del}
\end{table*}

\begin{table*}
\caption{Ablation studies of loss components based on SORD method.}
\setlength{\tabcolsep}{1mm}
\centering
\begin{tabular*}{1\linewidth}{@{\extracolsep{\fill}}cccccccccccccccccc}
\shline
\multicolumn{3}{c}{Loss Components} && \multicolumn{2}{c}{MORPH} && \multicolumn{2}{c}{DR} && \multicolumn{2}{c}{BUSI} && \multicolumn{2}{c}{HID} && \multicolumn{2}{c}{AADB} \\ 
\cline{1-3}  \cline{5-6}  \cline{8-9} \cline{11-12} \cline{14-15} \cline{17-18}
$\mathcal{L}_\mathrm{KL}$ & $\mathcal{L}_\mathrm{KL}^\mathrm{Dual}$ & $\mathcal{L}_{\text{Ent}}$ && MAE$\downarrow$ & CS (\%) $\uparrow$ && MAE$\downarrow$ & Acc. (\%) $\uparrow$ && MAE$\downarrow$ & Acc. (\%) $\uparrow$ && MAE$\downarrow$ & Acc. (\%) $\uparrow$ && SRCC$\uparrow$ & PLCC$\uparrow$\\

\midrule
\checkmark  & & & & 2.22      & 94.9 && 0.2527      & 82.7 && 0.2017 & 79.62 && 0.65 & 54.34 && 0.6689 & 0.6802 \\
& \checkmark  &   &         & 2.15      & \bf{95.6} && 0.2516      & 83.0 && 0.1923 & 80.89 && 0.61 & 55.47 && 0.6813 & 0.6898 \\
& \checkmark  & \checkmark && \bf{2.10} & \bf{95.6} && \bf{0.2504} & \bf{83.3} && \bf{0.1847} & \bf{82.20} && \bf{0.60} & \bf{55.69} && \bf{0.6832} & \bf{0.6993} \\
\shline
\end{tabular*}
\label{tab:KLvsDual}
\end{table*}
\begin{figure*}[t]
\centerline{\includegraphics[width=1.0\linewidth]{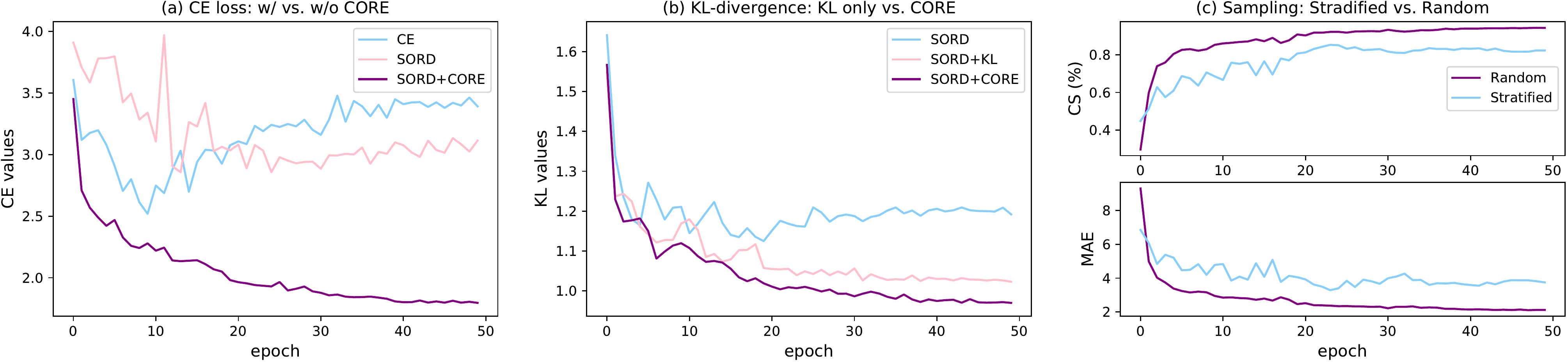}}
\caption{Ablation studies based on SORD baseline: (a) cross-entropy (CE), (b) KL-divergence, and (c) sampling strategies used during training.}
\label{fig:ablation}
\end{figure*}

\subsection{Ablation Studies}
\subsubsection{Effects of dual decomposition}
Table~\ref{tab:KLvsDual} shows the results of ablation studies \textit{w.r.t.} training with $\mathcal{L}_{\text{KL}}$ only, training with $\mathcal{L}_{\text{KL}}^{\text{Dual}}$, and training with $\mathcal{L}_{\text{KL}}^{\text{Dual}}+\mathcal{L}_{\text{Ent}}$. We can see that $\mathcal{L}_{\text{KL}}^{\text{Dual}}$ outperforms $\mathcal{L}_{\text{KL}}$ on the four applications. By minimizing Eq.~\eqref{eq:dual_kl_lambda}, $\lambda_{k}$ also serves as an ordinal prototype monitor of class $k$. Then, minimizing $\mathcal{L}_{\text{Ent}}$ helps the model avoid value collapse across different classes, which leads to  further performance gains. 
The variations in $\lambda_{k}$ are provided in Section 4 of the supplementary material.

\subsubsection{Comparisons between SORD and \methodname + SORD} In Fig.~\ref{fig:ablation}, we discuss differences in the CE loss and KL-divergence when the model is trained with different objectives. In Fig.~\ref{fig:ablation}(a), we compare the CE loss during training when the model is trained with the CE loss, SORD, and \methodname+SORD. We can see that the CE-trained model is more easily overfitted, while SORD does not experience overfitting, and is difficult to obtain lower CE loss. When combined with \methodname, SORD converges better. In Fig.~\ref{fig:ablation}(b), we compare the KL-divergence values using different methods. Note that the KL values in this experiment are computed directly between the OTDs of feature representations and ordinal labels. Compared with SORD only, SORD+KL shows slightly lower divergence values, which can be attributed to the optimization difficulty caused by the covariate shift between the input space and label space. However, \methodname enables a further decrease in KL-divergence. 
\begin{table}
\caption{Influence of different mini-batch sizes ($N_{B}$) with respect to \methodname based on SORD. The results are evaluated on MORPH.}
\centering
\begin{tabular*}{1\linewidth}{@{\extracolsep{\fill}}cccccc}
\shline
\multirow{2}{*}{\tabincell{c}{$N_{B}$}} & \multicolumn{2}{c}{SORD} && \multicolumn{2}{c}{SORD+\methodname}\\ \cline{2-3}  \cline{5-6}
 & MAE$\downarrow$ & CS (\%) $\uparrow$ && MAE$\downarrow$ & CS (\%) $\uparrow$ \\
\midrule
8   & 2.33 & 91.4 && 2.26 & 91.7 \\
16  & 2.32 & 91.7 && 2.30 & 91.8 \\
32  & 2.39 & 89.4 && \textbf{2.10} & \textbf{95.6} \\                   
64  & 2.37 & 90.0 && 2.28 & 90.0 \\ 
128 & 2.41 & 89.6 && 2.35 & 89.9 \\          
\shline
\end{tabular*}
\label{tab:batchsize}
\end{table}
\subsubsection{Sampling strategies for mini-batch}
We investigate two sampling strategies for mini-batch training with \methodname: 1) random sampling and 2) stratified sampling, \ie, $N_{K}=N_{B}$, where all the samples in a mini-batch have distinct ordinal labels. Fig.~\ref{fig:ablation}(c) shows that stratified sampling is inferior to random sampling. Note that stratified sampling leads to slight overfitting for SORD after 25 epochs. This can be explained from two aspects. First, as shown in Eq.~\eqref{eq:dual_kl_lambda}, stratified sampling invalidates the prototype constraint and results in the noncompactness of feature representations. Second, there are $N_{B}$ dual variables $\lambda$, each of which corresponds to a single sample such that $\lambda$ is not crucial for reflecting the importance of the corresponding class, \ie, the optimal solutions with prototypical constraint in Proposition~\ref{proposition:reparam} cannot be guaranteed. Therefore, stratified sampling poorly approximates the true distribution of the input space.

\subsubsection{Effects of different mini-batch sizes}
In this experiment, we evaluate the effects of varying the  values of $N_{B}$ on the MORPH dataset. Each OTD $\mat{Q}_{i,:}$ approaches the corresponding $\mat{P}_{i,:}$ through KL-divergence, averaged by $N_{B}$; hence, $N_{B}$ is independent of the number of classes in a given dataset. In Table~\ref{tab:batchsize}, we report the performances obtained by different values of $N_{B}$. We can see that the best value for \methodname is $N_{B}=32$, and SORD+\methodname nearly outperforms SORD for all $N_{B}$ values.

\subsubsection{Computational overhead}
\methodname is a plug-and-play component that is \emph{only} involved in the training stage. The training overhead is the computation of embedding distributions of features and ordinal labels and the KL-divergence in Eq.~\eqref{eq:dual_kl}, which has also been described in lines 8-10 of the pseudocode in Algorithm~\ref{alg:algorithm}. Quantitatively, we report the training time of one iteration averaged across an epoch. We can see from Table~\ref{tab:overhead} that the additional computational overhead is marginal compared to the significant performance improvements. However, at the inference stage, \methodname does not introduce any overhead to baseline methods.

\begin{table}[t]
\caption{Training time ($ms$) of one iteration averaged across one epoch on the MORPH dataset. Values in parentheses denote the additional computation overhead.}
\centering
\begin{tabular*}{1\linewidth}{@{\extracolsep{\fill}}cllll}
\shline
\methodname & SORD~\cite{SORD} & MV~\cite{Mean-variance} & Poisson~\cite{Unimodal} & POE~\cite{POE} \\
\midrule
           & 170            & 172            & 195           & 139            \\
\checkmark & 196$(+26)$ & 200$(+28)$ & 213$(+18)$ & 147$(+8)$ \\
\shline
\end{tabular*}
\label{tab:overhead}
\end{table}


\section{Conclusion}
\label{sec:conclusion}
In this paper, we proposed a consistent ordinal representation learning method for image ordinal estimation named \methodname. \methodname is motivated by ordinal consistency between feature representations and labels, and consistently improved the performance of existing ordinal regression methods on four different scenarios.
The visualization results and ablation studies further confirm the effectiveness of the proposed \methodname as well as its key components.
In summary, this paper offers new insights into the importance of the dominance of ordinal information in the embedding space for image ordinal estimation.

\balance


\vfill

\end{document}